\documentclass[a4paper,fleqn]{cas-scx}

\usepackage[authoryear,longnamesfirst]{natbib}

\usepackage{setspace}
\usepackage{lineno}

\usepackage{caption}
\usepackage{subcaption}

\def\tsc#1{\csdef{#1}{\textsc{\lowercase{#1}}\xspace}}
\tsc{WGM}
\tsc{QE}
\tsc{EP}
\tsc{PMS}
\tsc{BEC}
\tsc{DE}
\begin{document}

\let\WriteBookmarks\relax
\def\floatpagepagefraction{1}
\def\textpagefraction{.001}
\shorttitle{}
\shortauthors{Author et~al.}
%\begin{frontmatter}

\title [mode = title]{Visual Perception of Building and Household Vulnerability from Streets}                      

\author[1]{Chaofeng Wang}
\cormark[1]
\ead{{chaofeng.wang@ufl.edu}}
\address[1]{M.E. Rinker, Sr. School of Construction Management, College of Design, Construction and Planning, University of Florida, Florida, United States}

\author[2]{Sarah Elizabeth Antos}
\address[2]{World Bank Group, Washington, D.C., United States}

\author[2]{Jessica Grayson Gosling Goldsmith}

\author[2]{Luis Miguel Triveno}

\cortext[cor1]{*Corresponding author: Chaofeng Wang}

% % example of a footnote with no numbers
%\nonumnote{This note has no numbers. In this work we demonstrate $a_b$
%  the formation Y\_1 of a new type of polariton on the interface
%  between a cuprous oxide slab and a polystyrene micro-sphere placed
%  on the slab.
%  }

\begin{abstract}
In developing countries, building codes often are outdated or not enforced. As a result, a large portion of the housing stock is substandard and vulnerable to natural hazards and climate related events. 
Assessing housing quality is key to inform public policies and private investments. 
Standard assessment methods are typically carried out only on a sample / pilot basis due to its high costs or, when complete, tend to be obsolete due to the lack of compliance with recommended updating standards or not accessible to most users with the level of detail needed to take key policy or business decisions. 
Thus, we propose an evaluation framework that is cost-efficient for first capture and future updates, and is reliable at the block level. 
The framework complements existing work of using street view imagery combined with deep learning to automatically extract building information to assist the identification of housing characteristics.
We then check its potential for scalability and higher level reliability. For that purpose, we create an index, which synthesises the highest possible level of granularity of data at the housing unit and at the household level at the block level, and assess whether the predictions made by our model could be used to approximate vulnerability conditions with a lower budget and in selected areas.
Our results indicated that the predictions from the images are clearly correlated with the index.

\end{abstract}

%\includegraphics{figs/grabs.pdf}

\begin{keywords}

Household vulnerability \sep 
Deep learning \sep 
Disasters \sep 
Buildings \sep 
Street view images  

\end{keywords}

\maketitle

\begin{spacing}{2.0}

\section{Introduction}

Disastrous consequences of natural hazards may significantly impact a region on several levels, including casualties and damages to the housing stock and infrastructure, resulting in enormous economic losses at a value of USD \$2,908 billion in the past two decades \citep{Wallemacq2018}. As has been seen again and again, certain architectural constructs are vulnerable to certain types of natural hazards, such as earthquake and hurricanes. 
Today's understanding of the underlying elements driving the likelihood of the failure of a building is deeper than ever before. 
If the intensity of a natural hazard and the building's properties are known, the total risk of the building and the household could be evaluated \citep{stateoftheart2019}.

Although it is economically feasible to perform pre-disaster strengthening for large, 
heterogeneous building inventories for particular types of buildings \citep{kappos2008feasibility},
enforcing such mandates to upgrade structures has proved difficult. Pre-disaster upgrading of an existing building inventory of a city requires not only engineering and economic investigation, 
but also consideration of conflicting socioeconomic priorities. Take earthquake as an example, even in regions of high seismicity, where the direct witnesses of frequent earthquake events have paved the way to urban safety policy advances, the implementation of upgrading of seismically vulnerable buildings is slow - the government agencies need large-scale evaluation of the household vulnerability to make plans for retrofit programs. 
But the evaluation is a procedure that usually requires professional earthquake engineering knowledge and detailed collection of field data. 
These processes are costly and time-consuming. 
To reliably assess a structure's seismic vulnerability, in 1988, the Applied Technology Council (ATC) and the Federal Emergency Management Agency (FEMA) released the Rapid Visual Screening of Buildings for Potential Seismic Hazards: A Handbook \cite{fema154v1}. This handbook has been updated in several new versions along the years and it provides a systemic methodology to estimate the seismic safety of a buildings with minimum or no access to the inside of the buildings - mainly based on visually observing the buildings from the outside. It demonstrated that such vision-based screening procedures provide good results in general and can identify suspicious buildings for further risk investigation \citep{moseley2007pre}. Since it release, this method has been implemented in many countries and regions for rapid seismic evaluation of building stocks \citep{wallace2008seismic,karbassi2007adaptation,srikanth2010earthquake,saatcioglu2013seismic,perrone2015rapid,ploeger2016urban,ningthoujam2018rapid}. 
However, implementing such assessment procedures still requires an intensive use of professionals to conduct physical inspections. The traditional rapid visually screening of buildings \citep{fema154v1} is time consuming and expensive - the vulnerability screening for structure risks on a large scale, is not economically feasible.  
How to efficiently screen large inventories of these buildings is still problematic. 
To reduce the labor and time cost imposed on government agencies and, in addition, ease the economic burden on residents, 
a standard building and household vulnerability evaluation process for facilitating the automation of part of the investigative process is needed.

Street view imagery is an in-expensive data source. The collection of street view images requires minimum equipment - they can be captured using cameras mounted on moving objects such as vehicles and pedestrians. 
Its vast availability combined with recent breakthrough in computer vision (CV) algorithms has made it viable to rapidly screen large stocks of buildings. 
Because of it is easy to collect and can provide rich information of the street, significant attentions of this sought after resource has been drawn from researchers from multiple disciplines of the built environment. Therefore, a range of applications have been explored in recent years, such as housing price prediction \citep{bency2017beyond,law2018take,kang2021human} and neighborhood safety evaluation \citep{naik2014streetscore,liu2017place}, and many other aspects in urban planning \citep{ma2021measuring,samany2019automatic,taecharungroj2020big}. Since visual clues can manifest on building facades in images, it is possible to automatically identify vulnerable buildings using computer vision algorithms and street viewing techniques.

Meanwhile, breakthroughs in deep learning (DL) techniques and its successful application in computer vision have shown great potential for solving numerous engineering and social problems that were impossible to be solved before. For example, deep convolutional neural network (CNN), a core technique of DL, have found its applications in many areas such as civil, infrastructure, and natural hazards \citep{gao2018deep,wang2018damage,cha2018autonomous,wang2020autonomous,guofaccade,czerniawski2020automated}.

In this paper, we present a DL-based method that recognizes a building automatically from an image and further classify it regarding the construction type, material, use, and condition. The method is based on an instance segmentation technique. 
Further, to explore the link between the DL-extracted information and the the household vulnerability, and to check its potentials for scalability and higher level reliability, we create a index, K3, that is based on census data. The K3 index synthesises the highest possible level of granularity of data at the housing unit and at the household level at the block level, and assess whether the predictions made by our DL model could be used to approximate vulnerability conditions with a lower budget and in selected areas.
%Further, to eventually quantify the vulnerability of a house, we developed a index, K3, that is based on the census data at the block level. We performed a preliminary study to build the link between the detected building attributes and the K3 vulnerability index. 

\section{Vision-based Risk Evaluation}\label{sec:relatedwork}

Assessing risks of structures requires extensive examination of various information including the recognition of construction type and material, the deficiency in them, the condition of the maintenance, etc. 
To this end, the guidebook \textit{FEMA 154} \cite{fema154v1} and its updated versions \citet{fema154v2} and \citet{fema154v3} provide guidelines for assessing the performance of structures by employing a scoring scheme mainly based on the visual clues manifested at the exterior of buildings, without the need to access to the inner space of buildings.
The rationale under such visual-based assessment method is that 
the structure's performance largely depends on its construction type, material, maintenance, etc.

Although visual screening such as \textit{FEMA 154} has been broadly adopted, such screening can be expensive and prone to errors because it is extreme labor-intensive when gathering a vast number of data (i.e., images) of the buildings being investigated. Also the subjectiveness in human decisions can potentially lead to diverging interpretations and erroneous results.

To address this issue, this research presents an alternative procedure, which first collects street images from the region of interest using a ground vehicle and then employs a DL approach to recognize structures from these images. 

Street view data is relatively simple to collect and provides rich visual information of the road and buildings. Recent has proved that street view images are applicable to a variety of different studies. For example, the degree of urban public security can be calculated from visual cues perceived from numerous street view images. \cite{naik2014streetscore} found the machine-extracted visual presence of the urban environment can be used to measure the life quality of inhabitants in a neighborhood.
Similarly, \cite{gebru2017using} found the correlation between the street visual cues of cars and the demographic makeup by analyzing millions of street view images collected from multiple cities in the United States. In another study, \cite{law2018take} discovered a strong link between the housing price and the visual cues of building facades from the street. Street view images can also be combined with other sources of data to provide more details of the built environment, for example, \cite{kang2018building} utilized satellite and street view images to infer the function and occupancy type of buildings. All the aforementioned examples use DL-based methods to extract information from images.

Recently, the field of computer vision technology has experienced notable growth, which has benefited from the progress in machine learning. The research on deep convolutional neural networks (CNNs) has experienced notable growth, especially in the computer vision domain, such as object recognition/detection, semantic/instance segmentation, etc. 
For example, by applying DL to street view images, multiple features of buildings can be extracted directly \cite{yu2020rapid,wang2021automatic,wang2021machine}. This paper develops a DL-based system, in which the risk-related building attributes are automatically perceived from street view images and then correlated with household vulnerability.
The proposed system possesses many advantages when compared with traditional screening approaches, in terms of the consistency in the evaluation, cost efficiency, and scalability. 
%Despite the significant progress made in computer vision and image processing, applying deep neural network methods to evaluate building vulnerabilities from a raw image remains a challenging task because the visual features of a building in a complex street scene are local and subtle.

\section{Methodology} \label{Methodology}

\subsection{Overall workflow} \label{pipiline}

The objective of this research is to develop and validate an integrated workflow for automatic large-scale screening of building vulnerabilities at the city level. Components of the workflow include 1) Data Collection, 2) Annotation, 3) Model Training, and 4) Inference.
The core of the workflow is to successfully 
train instance segmentation models that are capable of identifying from street view images building instances and attributes. It is an challenging task because of the complex contexts in street view images. The model needs to be trained on a dataset that captures the variations in building facades so that the model can learn to identify building objects from unseen images and further classify the objects according to their different attributes.  \\

\noindent\textbf{Component 1: Collection of Data}\\
Street level images are first collected. Each image has its camera parameters and GPS coordinate recorded.  \\
\noindent\textbf{Component 2: Image Annotation}\\
For training the model, we created an annotated dataset, where the building objects in the collected street view images are annotated with bounding boxes. Each annotated building object is also labelled with the attributes, such as the use type, facade material, construction type, and the condition. In Section \ref{datacollection}, we describe the details of the image collection and annotation procedures. \\
\noindent\textbf{Component 3: Model Training}\\
Once the images are annotated, the next step is to train multiple models for instance segmentation. The details of the model can be found in \ref{model}.\\
\noindent\textbf{Component 4: Inference}\\
Once models are trained, they are then used for detecting buildings from other unseen images collected from the streets and predict the attributes of each detected building. \\
\noindent\textbf{Component 5: Geocoding}\\
In this step, the detected building attributes will be linked with building footprints. \\
\noindent\textbf{Component 6: Vulnerability Analysis}\\
The vulnerability of each household can be analyzed based on the information extracted from the the images.

\begin{figure*}[t]
\centering
\includegraphics[width=0.85\textwidth]{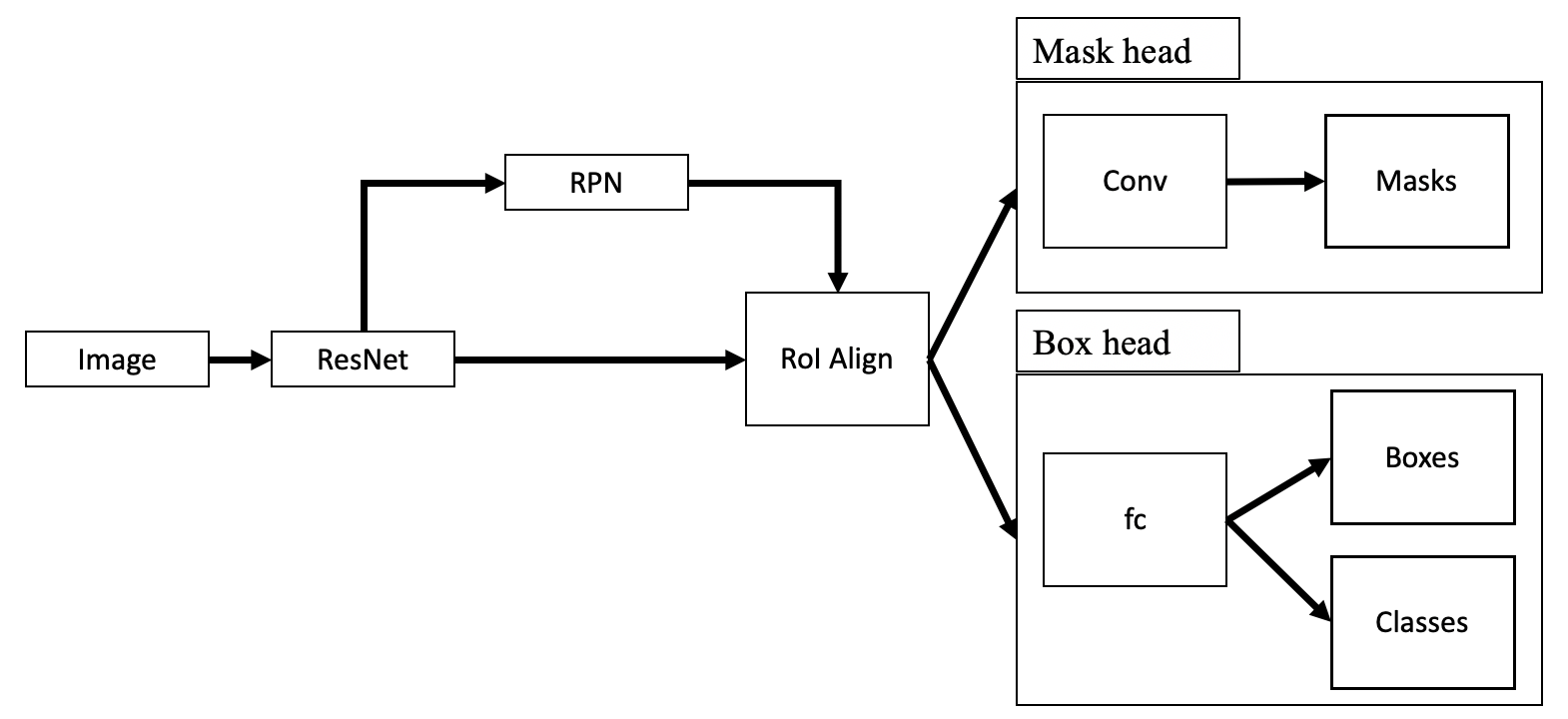}
\caption{Mask R-CNN \citep{he2017mask}}
\label{fig:mask-r-cnn}
\end{figure*}

\subsection{Model} \label{model}

An image classification model takes a given picture as input and returns the classification that determines whether the object of a specific class is displayed in the picture or not. In contrast, instance segmentation is about categorizing and grouping pixels in an image into predefined classes. 
In this study, we implement the instance segmentation technique in the workflow for 
information extraction from street view images, because it classifies the images at the pixel level and identifies the locations of building objects of specific categories.

There are various DL-based instance segmentation models that have been developed in recent years. The exact model we used for this task is based on the Mask R-CNN approach originally developed by \cite{he2017mask}. This approach is an extension of the Faster R-CNN algorithm \citep{ren2015faster}. 
Besides the original classification branch with bounding box regression, a new branch that can predict the segmentation masks is added to the head. We show the architecture of the Mask R-CNN in Fig.~\ref{fig:mask-r-cnn}. Details of the architecture can be found in its original paper. 

The backbone of the neural network is a ResNet, which is pretrained on the COCO dataset. 
A street view image is first fed into the backbone to generate a feature map, based on which a Region Proposal Network (RPN) can propose regions of interest (RoIs). Each RoI will enter head two branches for mask generation and bounding box prediction/classification.
The branch responsible for mask generation is a fully convolutional network that can predict the segmentation mask for each RoI at the pixel level. The other branch responsible for the bounding box is consisted of a set of fully-connected layers which performs the boudning box regression and softmax classification. The loss will be back-propagated to the network to update the weights. For each RoI, the loss can be calculated like this: $L=L_{cls}+L_{box}+L_{mask}$, in which \textit{L} represents the total loss; $L_{cls}$ represents the classification loss and $L_{box}$ represents the bounding box loss; and $L_{mask}$ represents the segmentation loss. The detailed definitions of $L_{cls}$, $L_{box}$ and $L_{mask}$ are referred to \citet{ren2015faster} and \citet{he2017mask}.

%\newpage

\subsection{Data Preparation} \label{datacollection}
\begin{figure*}[tb]
\centering
\includegraphics[width=0.65\textwidth]{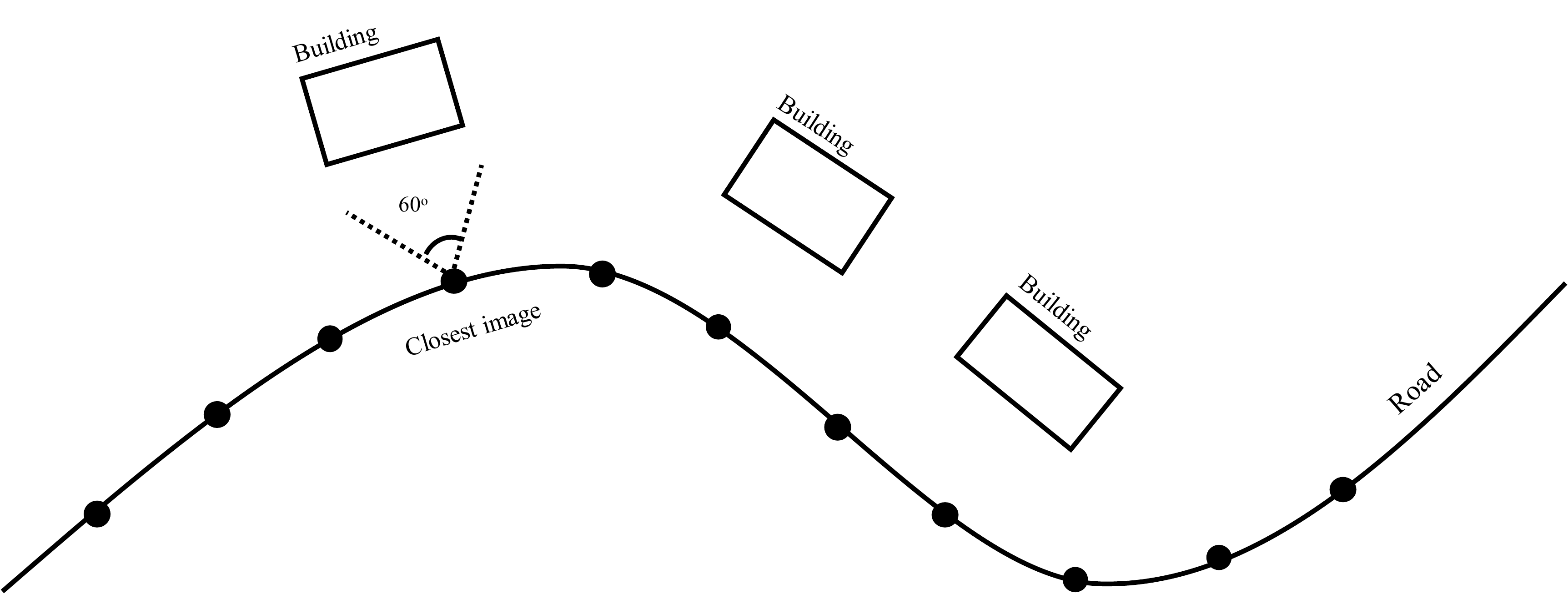}
\caption{Street view image acquisition}
\label{fig:svapi}
\end{figure*}

We collected all street view images using a 360 camera mounted on a ground vehicle.
As illustrated in \autoref{fig:svapi}, the vehicle drives through the streets with the cameras taking images at a constant speed. The GPS coordinate and camera parameters are recorded for each image at the moment the image is taken. The images selected for this research are those perpendicular to the travel direction.

\begin{figure*}[tb]
\centering
\includegraphics[width=0.8\textwidth]{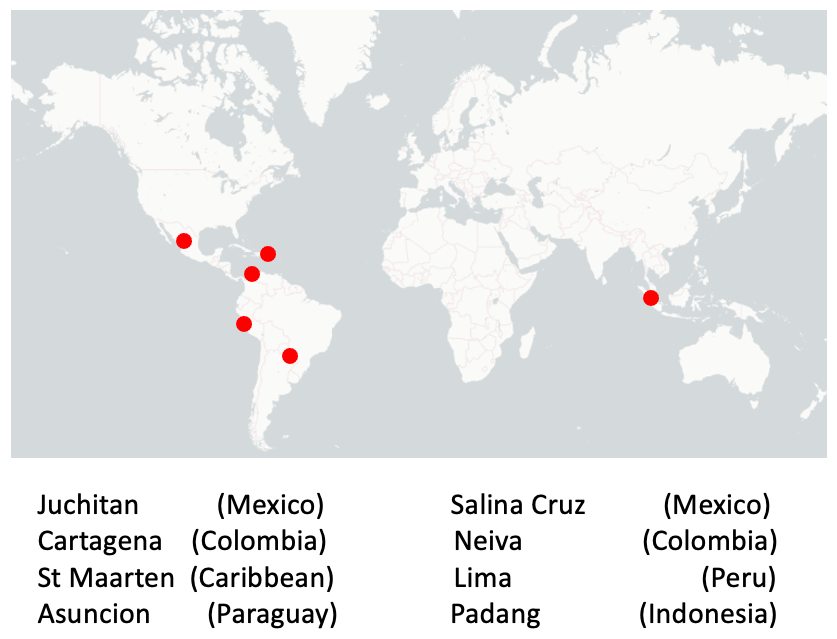}
\caption{Data source distribution}
\label{fig:cities}
\end{figure*} 

From all the street view images collected in 8 cities in both Latin America and Asia (as shown in Figure \ref{fig:cities}), a random subset is selected for annotation.
Building objects are annotated on the selected images using bounding boxes. 
Each bounding box is labelled with 4 tags: construction type, material, use, and condition. For each tag, the values are list in the Table \ref{tab:attr}.
A subset of 80\% of these annotations are used for training the model, the rest for validation.

\begin{table*}[!htbp]
    %\vspace{-10pt}
    \caption{Annotation tags}
    \centering
    \begin{center}
     \begin{tabular}{lll} 
     \toprule
     Tag Name & Value & Counts  \\ [1ex] 
             \midrule
        \vspace{.1in}  
     Construction type  & confined & 8,839 \\ 
        \vspace{.1in}  
       & unconfined & 1,642 \\ [1ex]
     \midrule
        \vspace{.1in}  

Material & plaster                          &          	76,584 \\
        \vspace{.1in}  
 & mix\_other\_unclear                  &           	19,446 \\
        \vspace{.1in}  
 & brick\_or\_concrete\_block           &     	16,613 \\
        \vspace{.1in}  
 & wood\_crude\_plank                    &           	2,303 \\
        \vspace{.1in}  
 & wood\_polished                      &            	1,097 \\
        \vspace{.1in}  
 & corrugated\_metal                   &             	566 \\
        \vspace{.1in}  
 & adobe                               &            	207 \\
        \vspace{.1in}  
 & stone\_with\_mud\_ashlar\_with\_lime\_or\_cement  &     	164 \\
        \vspace{.1in}  
 & container\_trailer                       &         	72 \\
        \vspace{.1in}  
 & plant\_material                        &           	46 \\
 \midrule
        \vspace{.1in}  

Use & residential     &   87,872 \\
\vspace{.1in}  
& non\_residential &   16,765 \\
\vspace{.1in}  
& mixed           &   12,461 \\
      \midrule
        \vspace{.1in}  
     Condition  & fair & 71,927  \\
        \vspace{.1in}  
     & poor &	28,768 \\
        \vspace{.1in}  
     & good &	16,403 \\
             \bottomrule 
    \end{tabular}
    \end{center}
    \label{tab:attr}
\end{table*}

\section{Model performance} \label{seg}

The aforementioned annotation dataset is used for training the Mask R-CNN model. 
This section shows the performance of the trained model.
A subset of annotated images that was not seen by the model during the training is used for calculating the performance. 
We estimate the performance based on Intersection over Union (IoU) greater than 75\%. IoU is calculated as the intersection of the prediction region and ground truth region divided by the union of prediction and ground truth regions.
For all IoU>75\% predictions, we show the accuracy and F1-score in Table

\begin{table*}[!htbp]
    %\vspace{-10pt}
    \caption{Accuracy and F1-score}
    \centering
    \begin{center}
     \begin{tabular}{lll} 
     \toprule
     Attribute	& Accuracy (\%) &	F1-score (\%)  \\ [1ex] 
             \midrule
        \vspace{.1in}  
Construction type &	98.97 &	98.97 \\
\vspace{.1in}  
Material &	97.08 &	93.30 \\
\vspace{.1in}  
Use &	94.62 &	88.31 \\
\vspace{.1in}  
Condition &	78.85 &	69.68 \\
             \bottomrule 
    \end{tabular}
    \end{center}
    \label{tab:accuracy}
\end{table*}

\begin{figure}
     \centering
     \begin{subfigure}[b]{0.24\textwidth}
         \centering
         \includegraphics[width=\textwidth]{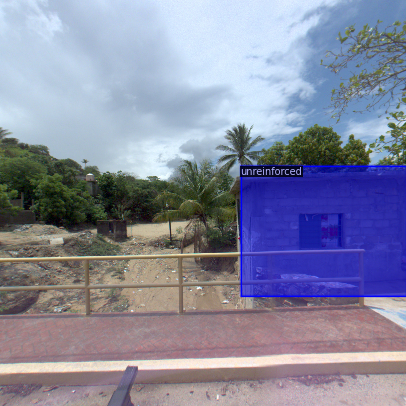}
         \caption{Construction annotation}
     \end{subfigure}
     \hfill
     \begin{subfigure}[b]{0.24\textwidth}
         \centering
         \includegraphics[width=\textwidth]{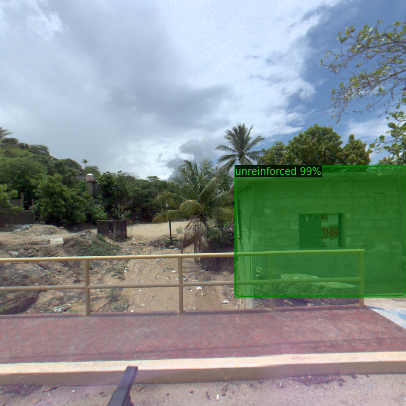}
         \caption{Construction segmentation}
     \end{subfigure}
     \hfill
     \begin{subfigure}[b]{0.24\textwidth}
         \centering
         \includegraphics[width=\textwidth]{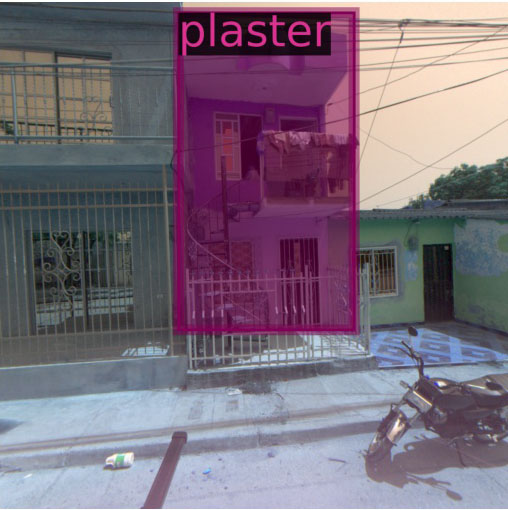}
         \caption{Material annotation}
     \end{subfigure}
     \hfill
     \begin{subfigure}[b]{0.24\textwidth}
         \centering
         \includegraphics[width=\textwidth]{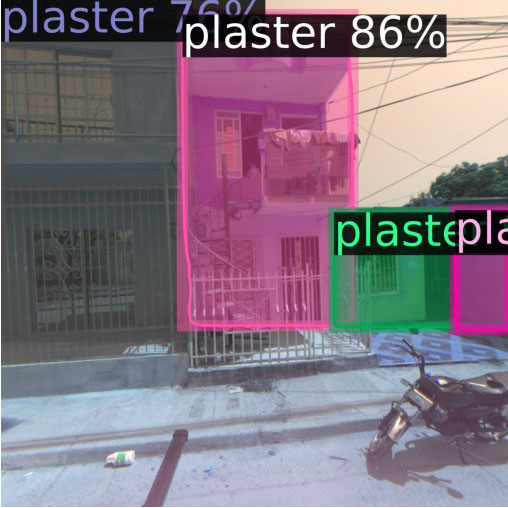}
         \caption{Material segmentation}
     \end{subfigure}
     \hfill
     \begin{subfigure}[b]{0.24\textwidth}
         \centering
         \includegraphics[width=\textwidth]{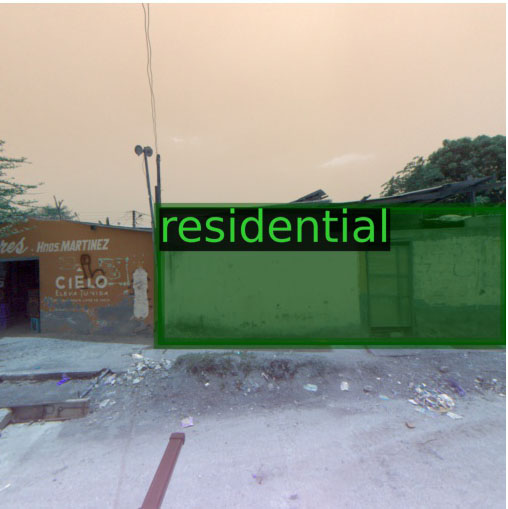}
         \caption{Use annotation}
     \end{subfigure}
     \hfill
     \begin{subfigure}[b]{0.24\textwidth}
         \centering
         \includegraphics[width=\textwidth]{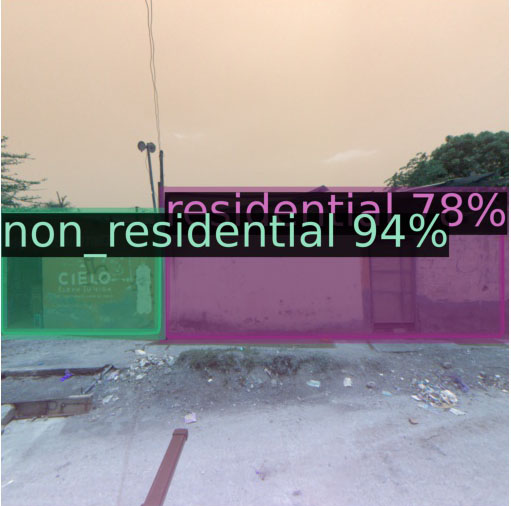}
         \caption{Use segmentation}
     \end{subfigure}
     \hfill
     \begin{subfigure}[b]{0.24\textwidth}
         \centering
         \includegraphics[width=\textwidth]{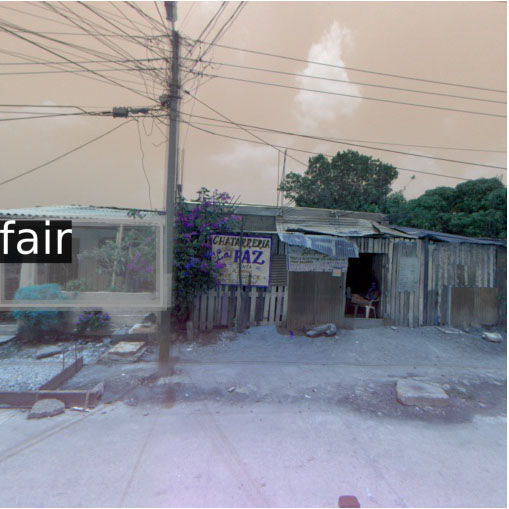}
         \caption{Condition annotation}
     \end{subfigure}
     \hfill
     \begin{subfigure}[b]{0.24\textwidth}
         \centering
         \includegraphics[width=\textwidth]{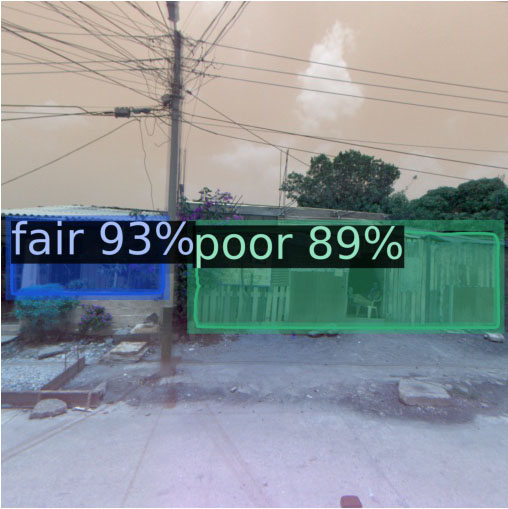}
         \caption{Condition segmentation}
     \end{subfigure}
\caption{Segmentation results compared with ground truth annotations}
\label{fig:result_examples}
\end{figure}

A few prediction examples are presented in Figure \ref{fig:result_examples}, followed by a discussion on the performance.

\section{Household Vulnerability}

To test the usefulness / utility of the automated predictions, we developed the K3 index for quantifying the household vulnerability.
K3 is a proxy for a ground-truth testing. It incorporates granular data that is generally unavailable. 
The index is calculated based on 27 parameters that we obtained from census data.
First, for each household, all 27 variables are standardized in values from 0 to 1. 
Then, Gower's similarity index are calculated, based on which we performed the clustering analysis to group the households into three groups. So every household gets classified and assigned a number according to its group (1, 2 or 3). 
For a whole census block, we average the numbers of all households.
To make sure that all the clusters are really “different”, an ANOVA analysis is conducted.
The final value for each block is called the K3 index. Lower K3 value means the household is more vulnerable from the socio-economic point of view.

\begin{table*}[!htbp]
    %\vspace{-10pt}
    \caption{Household parameters for evaluating K3}
    \centering
    \begin{center}
     \begin{tabular}{lll} 
     \toprule
     Category &	Parameter  \\ [1ex] 
             \midrule
        \vspace{.1in}  
Housing unit (Yes/No) & More than one household per housing unit \\
\vspace{.1in}  
& Walls made of industrial materials? \\
\vspace{.1in}  
& Floor made of industrial materials? \\
\vspace{.1in}  
& Access to electricity? \\
\vspace{.1in} 
& Access to water? \\
\vspace{.1in}  
& Access to sewerage? \\
\vspace{.1in}  
& Access to natural gas? \\
\vspace{.1in}  
& Access to waste collection services? \\
\vspace{.1in}  
& Waste collection more than 3 times a week? \\
\vspace{.1in}  
& Access to internet connection (fixed or mobile)? \\
\vspace{.1in}  
& WC connected to sewage network? \\
\vspace{.1in}  
& At least one room in the house is a bedroom? \\
\vspace{.1in}  
& Independent room for the kitchen? \\
\vspace{.1in}  
& Kitchen connected to the water network? \\
\vspace{.1in}  
& Less than 3 persons per bedroom? \\
\vspace{.1in}  
Household Members (Percentages) & Members that are men.  \\
\vspace{.1in}  
& Members order than 64. \\
\vspace{.1in}  
& Households without members of indigenous origin. \\
\vspace{.1in}  
& Members that live in the municipality they were born. \\
\vspace{.1in}  
& Members that did not get sick. \\
\vspace{.1in}  
& Members without any disability. \\
\vspace{.1in}  
& Members that are not illiterate. \\
\vspace{.1in}  
& Members above 24 years old with college or higher education. \\
\vspace{.1in}  
& Members between the ages of 15 and 64 that are working. \\
\vspace{.1in}  
& Members that are married or in another partnership arrangement. \\
\vspace{.1in}  
& Women that have children \\
             \bottomrule 
    \end{tabular}
    \end{center}
    \label{tab:k3}
\end{table*}

\begin{figure}
     \centering
     \begin{subfigure}[b]{0.48\textwidth}
         \centering
         \includegraphics[width=\textwidth]{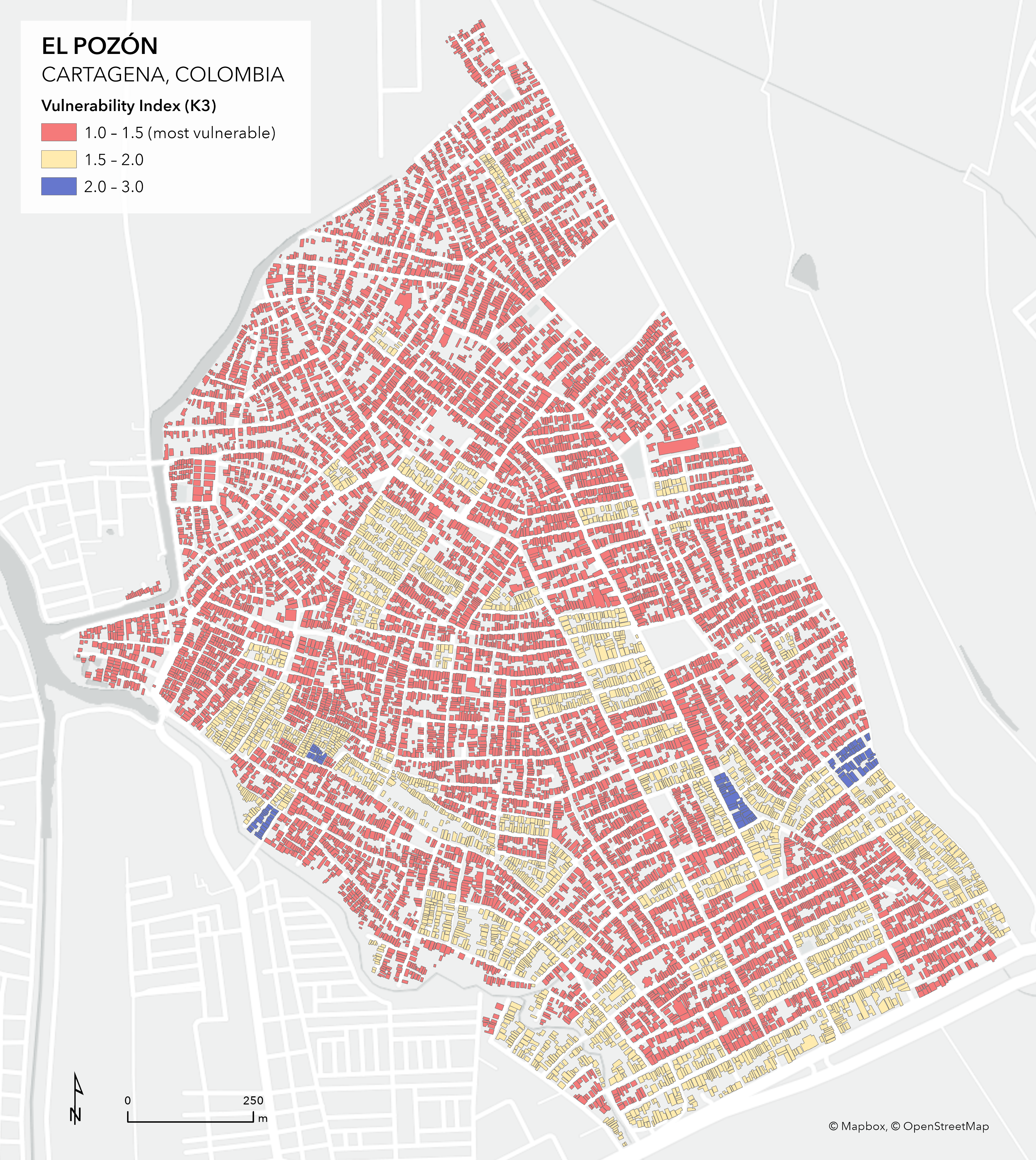}
         \caption{El Pozón K3 map}
     \end{subfigure}
     \hfill
     \begin{subfigure}[b]{0.48\textwidth}
         \centering
         \includegraphics[width=\textwidth]{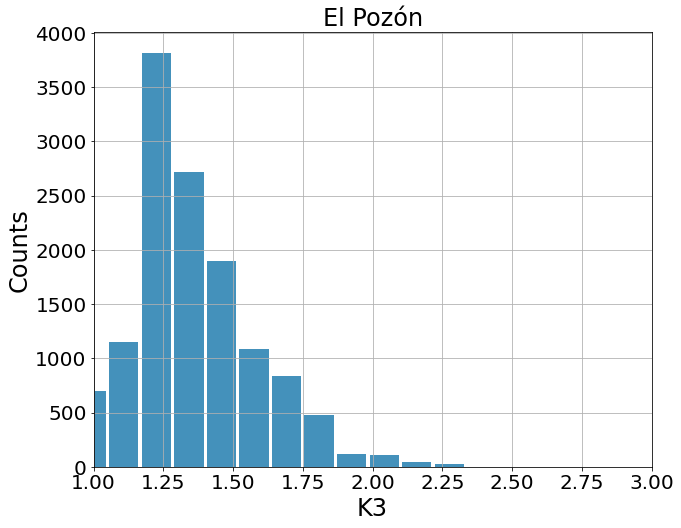}
         \caption{El Pozón K3 histogram}
     \end{subfigure}
     \hfill
     \begin{subfigure}[b]{0.48\textwidth}
         \centering
         \includegraphics[width=\textwidth]{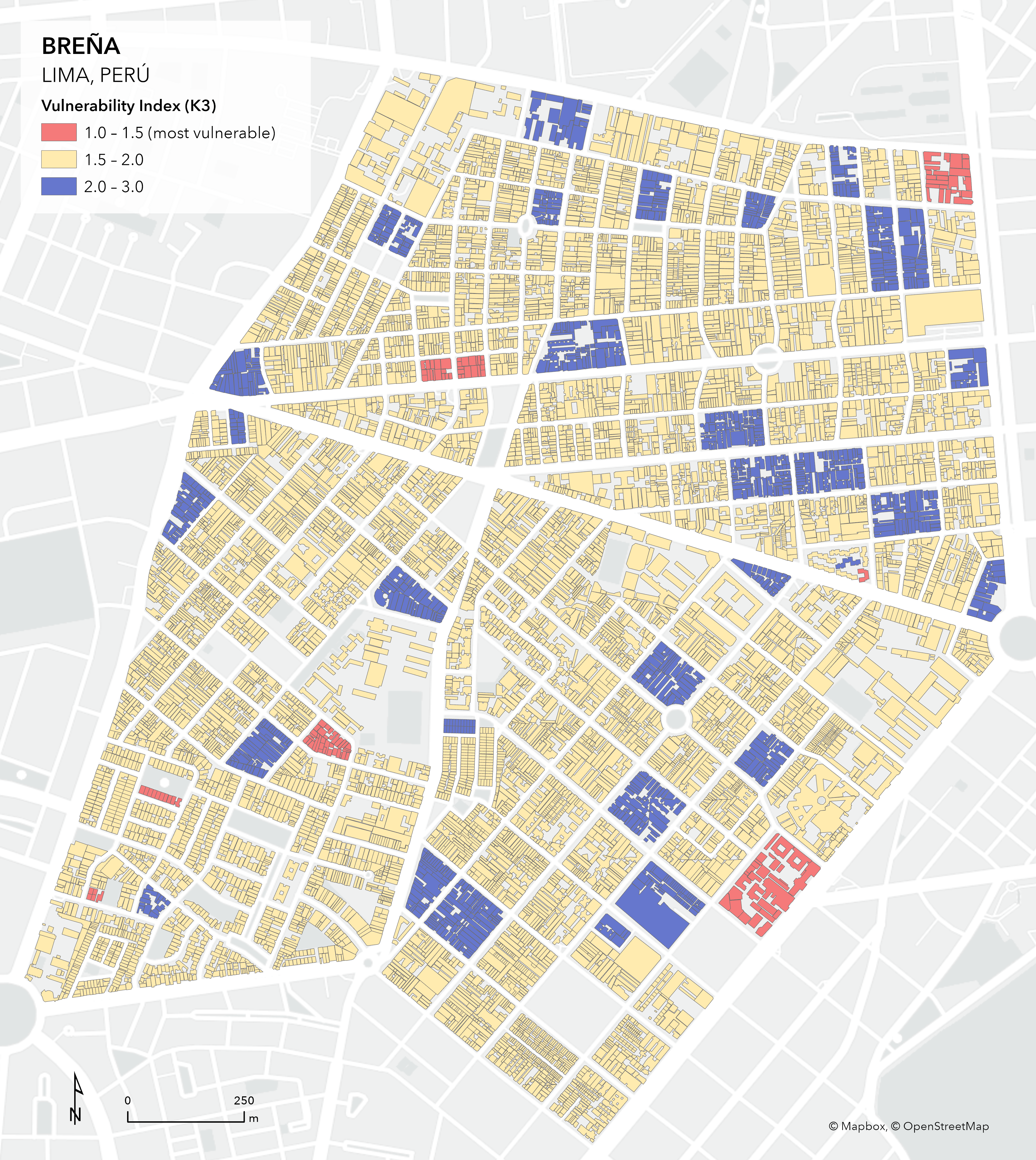}
         \caption{Breña K3 map}
     \end{subfigure}
     \hfill
     \begin{subfigure}[b]{0.48\textwidth}
         \centering
         \includegraphics[width=\textwidth]{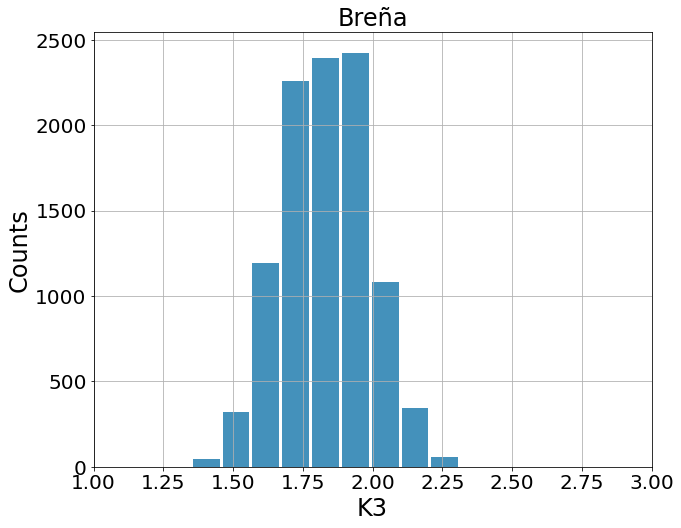}
         \caption{Breña K3 histogram}
     \end{subfigure}

\caption{K3 distributions}
\label{fig:K3_map}
\end{figure}

To explore the relationship between the predicted housing characteristics and the socio-economic status of those living in them, two neighborhoods were viewed with distinct qualities. 
The first is Breña, a densely-populated neighborhood near the historical city center of Lima, Peru. Breña (3.2 sq km) is a middle- to low-income neighborhood that was formally established in the mid-twentieth century. In contrast, the second is El Pozón (2.4 sq km), an informally established neighborhood of low-income residents living on the periphery of Cartagena, Colombia. Their K3 map and the histogram are plotted in Figure \ref{fig:K3_map}. From the figures, we can see most households in El Pozón have a K3 value less than 1.5 (more vulnerable), while most households in Breña are greater than 1.5 (less vulnerable). 
This means our K3 index is capable of capture the distinctions between households.

\begin{figure*}[t]
\centering
\includegraphics[width=0.5\textwidth]{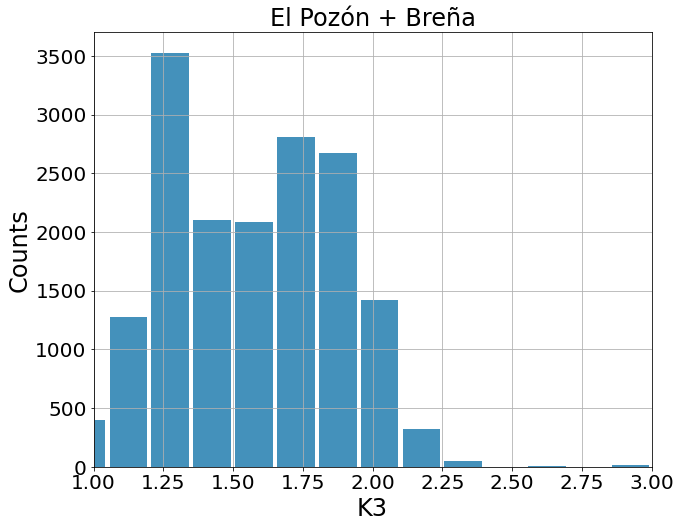}
\caption{Histogram of K3 (El Pozon + Brena)}
\label{fig:hist}
\end{figure*}

\begin{figure*}[t]
\centering
\includegraphics[width=0.5\textwidth]{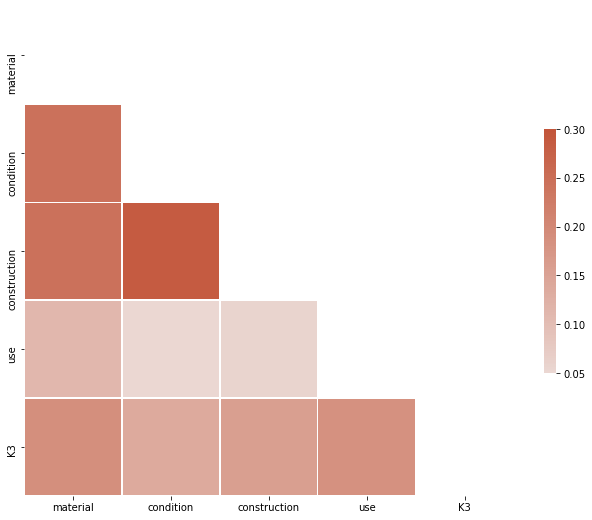}
\caption{Correlation Matrix (El Pozon + Brena)}
\label{fig:correlation}
\end{figure*}

We then combined the K3 data for two neighborhoods. 
The histogram of the combined K3 data is presented in Figure \ref{fig:hist}. 
For each household, we use the segmentation model to infer the building's attributes (construction type, material, use, and condition) from the street view images. We then investigated the correlation between the DL-predicted information and the K3 index. The correlation matrix between all parameters are presented in Figure \ref{fig:correlation}, showing that these variables correlates with each other. 

\begin{figure}
     \centering
     \begin{subfigure}[b]{0.48\textwidth}
         \centering
         \includegraphics[width=\textwidth]{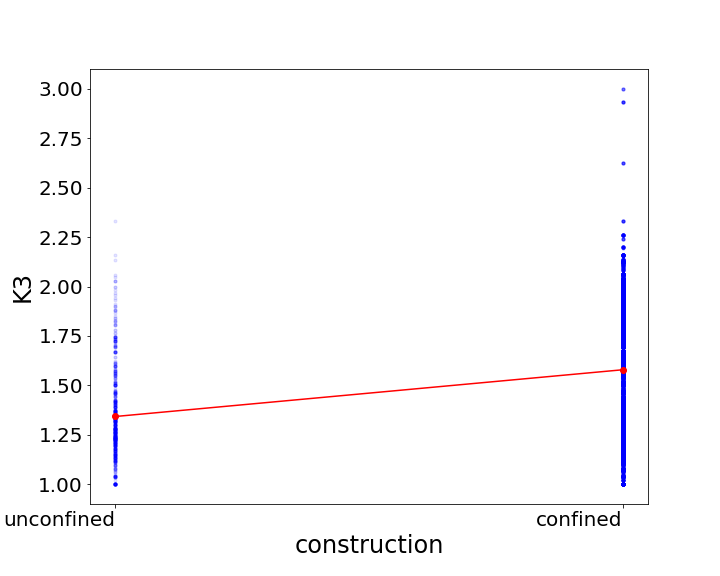}
         \caption{Construction type \\\textcolor{white}{=wood (polished), Mat2=wood (crude plank), Mat3=corrugated metal, Mat4=brick or concrete block, Mat5=plaster, Mat6=mix, Mat5=plaster, Mat6=mix}}
     \end{subfigure}
     \hfill
     \begin{subfigure}[b]{0.48\textwidth}
         \centering
         \includegraphics[width=\textwidth]{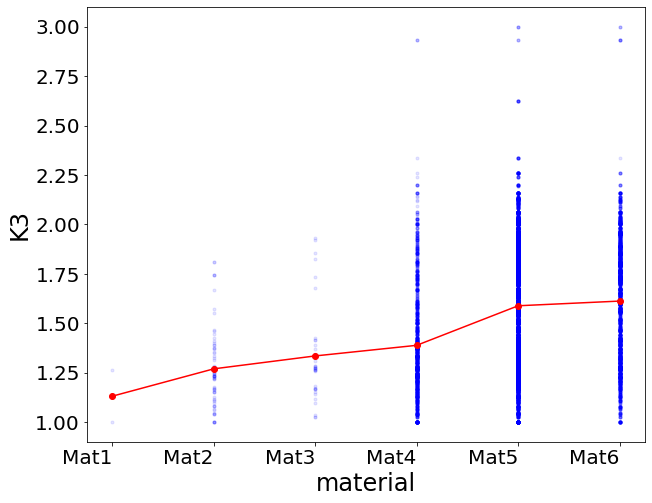}
         \caption{Material: Mat1=wood (polished), Mat2=wood (crude plank), Mat3=corrugated metal, Mat4=brick or concrete block, Mat5=plaster, Mat6=mix}
     \end{subfigure}
     \hfill
     \begin{subfigure}[b]{0.48\textwidth}
         \centering
         \includegraphics[width=\textwidth]{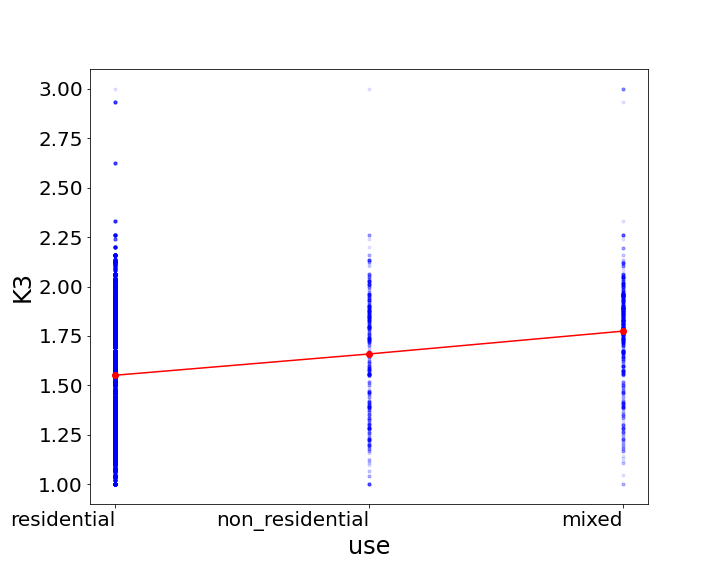}
         \caption{Use}
     \end{subfigure}
     \hfill
     \begin{subfigure}[b]{0.48\textwidth}
         \centering
         \includegraphics[width=\textwidth]{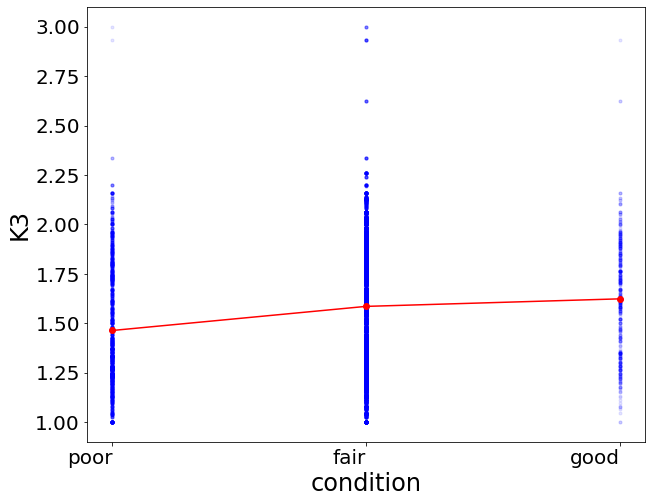}
         \caption{Condition}
     \end{subfigure}

\caption{Predictions Versus K3 Index}
\label{fig:K3}
\end{figure}

To further investigate the correlation between the predictions from images with the K3 index, 
we present the scatter plots and the trending line in Figure \ref{fig:K3}. 
All these figures proves again that the predictions are correlated with the K3 index.
Regarding the construction type, it shows the unconfined buildings have lower K3 values, indicating more vulnerable. 
For building materials, mix, plaster, brick or concrete block have the highest K3 values, while wood and corrugated metal buildings are found with lower K3 values. 
Regarding the use of the building, non residential buildings have higher K3 values than residential, while the mixed building type has the highest.
For the condition prediction, the correlation is also very clear and reasonable: poor < fair < good.
Base on these observations, we believe that the prediction from street view images and the K3 derived from census data agree with each other very well. Thus, it could be possible to estimate the  vulnerability of the building and household using deep learning and street view images.

\section{Discussion} \label{discussion}

Increasing natural hazard causes disastrous consequences that significantly impact the built environment.
It is essential to learn the vulnerability of large stock of buildings, especially in developing countries, because assessing housing vulnerability is key to inform public policies (e.g., housing subsidies, urban upgrading, social cash transfers, etc.) and private investments (e.g., real estate, credit, insurance, education, health and entertainment services, etc.). This study demonstrates that deep learning-based image segmentation can be used to help identify house attributes from street view imagery. 
The performance of the trained model is promising: the construction type accuracy is 98.97\%, the material type accuracy is 97.08\%, the use class accuracy is 94.62\%, while the condition classification accuracy is 78.85\%. 

It should be noted that the annotation dataset used for training is imbalanced. 
For example, in the material labels, there are three major classes (plaster/mix\_other\_unclear/brick\_or\_concrete\_block) that are dominant, while the left minor classes have much less labels ranging from only 46 to 2,000. 
Such imbalances could cause difficulties for the model to recognize the minor classes. 
Future improvements are possible with more labels in those classes. However, it should also be noted that these minor classes are not common, therefore they are always hard to find.

Among the four building attributes, the condition prediction has the lowest performance. 
There are two possible reasons. First, the training dataset is imbalanced. There are much fewer labels for the 'good' class. Second, it is hard to distinguish 'good' and 'fair', even for trained labelers.
We investigated the annotated images and found bias in the labels regarding the two classes. 
The bias and noise could possibly be one of the causes, therefore the model might be improved by eliminating those in future studies. 

The K3 index has the potential to be used for quantifying the vulnerability of a household.
It should be noted that we found the clear correlation between the street view predictions and the K3 index.
This implies that it is possible to evaluate the household vulnerability directly from street view images. With the deep learning model, the evaluation procedure is scalable.

\section{Conclusion} \label{conclusion}

This study presents an automated workflow to collect building attributes from street view images with deep learning-based instance segmentation method.  
The model can detect four building attributes with high accuracy: construction type (98.97\%), material (97.08\%), use (94.62\%), and condition (78.85\%). The model is broadly applicable to regions that have similar geographical street views, indicating the similar construction types, development, and economy levels. 

We demonstrated that a simple census data index, K3, can be can be used for quantifying the vulnerability of a household: financially robust households have higher K3 values, while households with lower K3 values are more vulnerable. 

Applying the developed segmentation model and K3 model to two neighborhoods,  Breña, Peru and El Pozón, Colombia, we found a clear correlation between these two sources. 
Therefore, we believe it is possible to develop a deep learning-based automatic system to rapid evaluate household vulnerabilities from street view images. 

This work seems to be the first study to use deep learning-based image analysis for household vulnerability study. The present approach aims at scalability and higher level reliability - it provides an automated and inexpensive method for large-scale regional examinations of vulnerability at the household level. The method requires minimum interactions, providing flexibility that enables implementations even during a period like COVID-19. It overcomes the difficulties in traditional standard assessments that are expensive or dependent of either slowly-developed datasets such as census or third-party datasets inaccessible to most users with the level of detail needed to take key policy or business decisions.

%\section{Acknowledgments}

%The methodology applied to create the census-based K3 index was an original idea from Farid Matuk. The images, labels and algorithms used in this paper were generated by the team of the World Bank's Global Program for Resilient Housing. 

\end{spacing}

\newpage

\bibliographystyle{cas-model2-names}

\bibliography{cas-refs}


\end{document}